\documentclass[journal,comsoc]{IEEEtran}
\usepackage[T1]{fontenc}% optional T1 font encoding
\usepackage{makecell}

\usepackage{graphicx}   %插入图片的宏包
\usepackage{float}      %设置图片浮动位置的宏包
\usepackage{subfigure}  %插入多图时用子图显示的宏包
\usepackage{booktabs}
\usepackage{multirow}
\usepackage{amssymb}
\usepackage{bm}
\usepackage{color, xcolor}
\usepackage{amsmath}
\usepackage{algorithm2e}
\RestyleAlgo{ruled}
\usepackage{enumitem}
\usepackage[pagebackref=true, colorlinks]{hyperref}
\usepackage{tabularray}
\usepackage{lipsum}
\usepackage[export]{adjustbox}
\usepackage{anyfontsize}

\newlength{\bibitemsep}
\newlength{\bibparskip}
\let\oldthebibliography\thebibliography
\renewcommand\thebibliography[1]{%
  \oldthebibliography{#1}%
  \setlength{\parskip}{\bibitemsep}%
  \setlength{\itemsep}{\bibparskip}%
}

\usepackage[cmintegrals]{newtxmath}
\newcommand{\etal}{\textit{et~al.}}

\begin{document}

%\title{Self-Supervised Adapter Learning Bridges the Gap: Foundation Models for Cross-View geo-localization}
\title{Unsupervised Multi-view UAV Image Geo-localization via Iterative Rendering}

\author {
Haoyuan~Li,
\and Chang~Xu, 
\and Wen~Yang, %~\IEEEmembership{Senior~Member,~IEEE,}
\and Li~Mi,
\and Huai~Yu,~%~\IEEEmembership{Member,~IEEE,}
\and Haijian~Zhang%,~\IEEEmembership{Senior~Member,~IEEE}

% \thanks{This work was supported in part by the National Natural Science Foundation of China (NSFC) under Grant 62271355 and NSFC Regional Innovation and Development Joint Fund (No. U22A2010).}
\thanks{H. Li, W. Yang, H. Yu, and H. Zhang are with the School of Electronic Information, Wuhan University, Wuhan, 430072 China. C. Xu and L. Mi are with École Polytechnique Fédérale de Lausanne. \emph{E-mail: \{lihaoyuan, yangwen, yuhuai, haijian.zhang \}@whu.edu.cn, \{chang.xu, li.mi\}@epfl.ch}.}
}

\maketitle

\begin{abstract}
Unmanned Aerial Vehicle (UAV) Cross-View Geo-Localization (CVGL) presents significant challenges due to the view discrepancy between oblique UAV images and overhead satellite images. 
Existing methods heavily rely on the supervision of labeled datasets to extract viewpoint-invariant features for cross-view retrieval. 
However, these methods have expensive training costs and tend to overfit the region-specific cues, showing limited generalizability to new regions. 
To overcome this issue, we propose an unsupervised solution that lifts the scene representation to 3d space from UAV observations for satellite image generation, providing robust representation against view distortion.
By generating orthogonal images that closely resemble satellite views, our method reduces view discrepancies in feature representation and mitigates shortcuts in region-specific image pairing. 
To further align the rendered image’s perspective with the real one,
we design an iterative camera pose updating mechanism that progressively modulates the rendered query image with potential satellite targets, eliminating spatial offsets relative to the reference images. Additionally, this iterative refinement strategy enhances cross-view feature invariance through view-consistent fusion across iterations. 
As such, our unsupervised paradigm naturally avoids the problem of region-specific overfitting, enabling generic CVGL for UAV images without feature fine-tuning or data-driven training. 
Experiments on the University-1652 and SUES-200 datasets demonstrate that our approach significantly improves geo-localization accuracy while maintaining robustness across diverse regions. Notably, without model fine-tuning or paired training, our method achieves competitive performance with recent supervised methods. 

\end{abstract}

% Note that keywords are not normally used for preview papers.
\begin{IEEEkeywords}
Cross-view Geo-localization, Foundation Model, Gaussian Splatting, Unmanned Aerial Vehicle
\end{IEEEkeywords}

\IEEEpeerreviewmaketitle

\section{Introduction}

\IEEEPARstart{A}ccurately determining the geo-location of a scene from drone images in GPS-denied environments is crucial for various applications, including autonomous navigation~\cite{xue2022terrain}, disaster response~\cite{Firmansyah2024Improving}, and  surveillance~\cite{kontitsis2004uav}. This task is commonly formulated as a retrieval problem, requiring the model to find the corresponding geo-tagged satellite images for drone-view queries~\cite{ren2023hashing, zhu2023sues}. This process, referred to as Unmanned Aerial Vehicle (UAV) Cross-View Visual Geo-Localization (CVGL), is particularly challenging due to the substantial view discrepancy and style variations between the orthogonal satellite imagery and the oblique perspectives captured by drones.

\begin{figure}[t]
    \centering
    \includegraphics[width=0.45\textwidth]{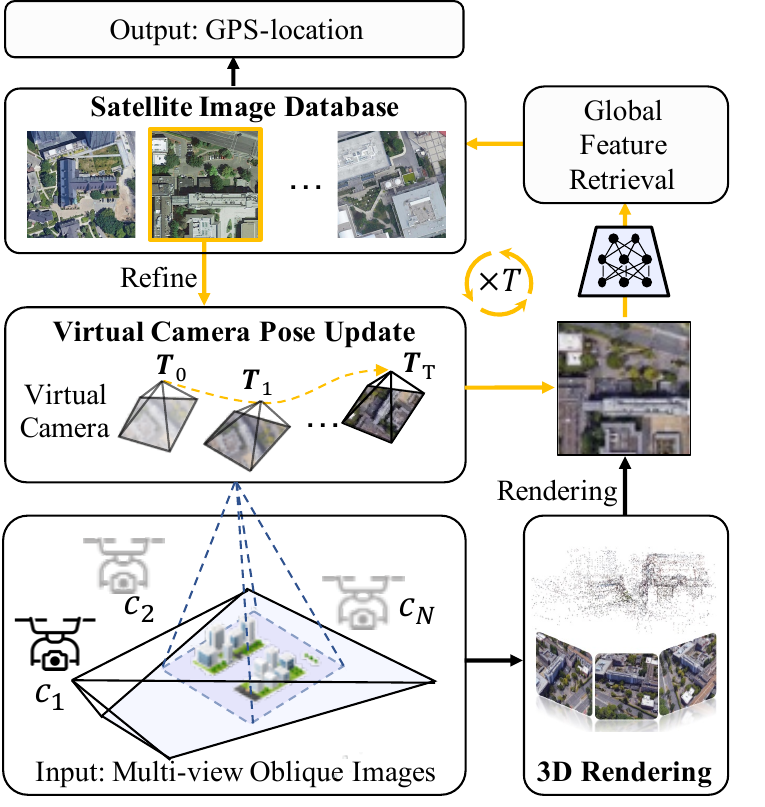}
    \centering
    \caption{\textbf{Illustration of the proposed scene rendering approach for cross-view geo-localization.} We propose a multi-view rendering image regression and retrieval approach that uses multiple UAV-captured views of a scene to predict its geo-location by retrieving matching satellite images from a database. First, we establish a 3D field representation of the query scene from multiple oblique images. Then, we iteratively render the virtual satellite view to align with the real satellite image from the database (yellow box). For notational convenience, we use $\bm{T}$ to represent both rotation $\bm{R}$ and translations $\bm{t}$ of camera poses.}
    \label{fig:outline}
\end{figure}

In cross-view retrieval tasks, a significant challenge for feature-based methods lies in establishing consistent feature representation across different perspectives. 
Current methods~\cite{shen2024mccg, Deuser_2023_ICCV, haas2023learning, xia2024enhancing} successfully employ deep neural networks to extract view-consistent cues (\textit{e.g.}, color, texture, shape) under paired image supervision.
However, depth information, which is a robust representation across scenarios, is lost 
during feature alignment in the 2D space. 
%which can lead to overfitting when trained on region-specific image pairs. 
As a result, models supervised by region-specific image pairs tend to degrade significantly in new regions with varying environmental conditions, for example, cross-city~\cite{yang2021cross, wang2024multiple} or planet-scale geo-localization~\cite{haas2023learning}. 
While recent works~\cite{li2024unleashing} and ~\cite{li2024learning} have sought to generalize models to new, unlabeled regions, these approaches still rely on pre-training with annotated data from source regions~\cite{xia2024adapting} or fine-tuning on target regions using pseudo-labels~\cite{li2024unleashing,li2024learning,xia2024adapting}. 
Such methods remain vulnerable to overfitting the target distribution, limiting their ability to generalize across unseen data sources during pretraining or fine-tuning. Therefore, this raises an interesting question: \textbf{Can we develop a drone-based CVGL method that establishes consistent scene representation without relying on network re-training in the 2D space?
}

To answer this question, we propose a novel approach that recovers scene structures and lifts the query into 3D space, providing a robust representation to retrieve satellite views without training. Our method leverages query observations to render satellite correspondences via a 3D neural field, achieving a region-agnostic representation for retrieval, as shown in Fig.~\ref{fig:outline}. Specifically, a 3D Gaussian Splatting (3DGS) model is employed to represent the 3D scene from drone-view observations in a neural field. This 3D scene is a shared space across drone-satellite views, enabling the generation of high-quality satellite views that narrow the view discrepancy.   

Despite the benefits of 3D neural representation, directly rendered images from 3DGS can still exhibit spatial shifts, scale differences, and rotations relative to true satellite images. These discrepancies lead to misalignments in perspectives and fine-grained details. Therefore, we further introduce an iterative update mechanism to refine the virtual camera poses. This mechanism leverages the spatial relationships of retrieved satellite candidates, progressively aligning the rendered image with potential matches. Finally, the rendered and candidate images are combined into a robust scene representation through a fusion module that incorporates view consistency between the features. Through the 3D neural scene representation and feature refinement, our approach significantly improves geo-localization performance without the need for network fine-tuning or supervised training.

We evaluate our pipeline on two standard multi-view drone image geo-localization benchmarks, University-1652~\cite{zheng2020university}, and the SUES-200~\cite{zhu2023sues} dataset. Both datasets contain multiple drone images of each ground target and corresponding geo-tagged satellite patches.  The experiments show that our novel unsupervised pipeline outperforms the recent SOTA zero-shot method, as well as achieves competitive performance with the task-specified supervised methods.

In summary, our contributions are outlined below:
\begin{itemize}	
\item We propose an unsupervised approach for multi-view UAV image geo-localization without network fine-tuning, leveraging 3D radiance field rendering to produce robust cross-view feature representation.
%We propose the first unsupervised paradigm for multi-view UAV image geo-localization without network fine-tuning, leveraging scene rendering and foundation models to produce robust cross-view feature representation.

\item We introduce an iterative camera pose update process that simultaneously aligns the rendered query to the satellite image and refines the retrieval results.

\item We validate the effectiveness of our approach through extensive experiments, showing its generalization to different regions and datasets. 
\end{itemize}

The rest of the paper is organized as follows: Section~\ref{sec:related_work} reviews the related work, focusing on cross-view geo-localization, novel view synthesis techniques, and multi-view techniques for geo-localization. In Section~\ref{sec:method}, we detail our proposed pipeline, including the method overview, the orthogonal Gaussian scene rendering, and the iterative camera pose update process. Section~\ref{sec:experiment} presents our experimental results and evaluates the performance of the pipeline in various datasets, with an in-depth analysis of its generalization and robustness. Finally, Section~\ref{sec:conclusion} provides concluding remarks.

\section{Related Work}
\label{sec:related_work}
In this section, we review the recent progress highly relevant to this work, including cross-view image geo-localization techniques, novel view synthesis for geo-localization, and the multi-view reconstruction techniques adaptable for geo-localization.

\subsection{Cross-view Place Geo-localization}

Traditional UAV geo-localization methods typically rely on hand-crafted feature matching~\cite{lowe2004distinctive} and rough GPS data for pose estimation~\cite{shetty2019uav}. Recent approaches aim to estimate global geographical coordinates over larger regions by matching query images to geo-tagged satellite patches, framing the task as an image retrieval problem. Numerous works~\cite{guo2022soft, hu2018cvm, regmi2019bridging} focus on retrieval between ground-level panoramas (e.g., street views) and satellite images. Despite the significant visual differences in cross-view images,  the ground-level field of view can be well-assigned with a single, identifiable satellite patch~\cite{zhu2021vigor}. In contrast, UAV images, captured from oblique angles, typically span multiple satellite patches, which introduces uncertainty and challenges in establishing accurate matching relationships~\cite{zheng2023uavs}. 

To address this uncertainty in UAV geo-localization, Zheng~\etal~\cite{zheng2020university} introduce the first multi-view multi-source dataset specifically designed for drone-to-satellite geo-localization. Many other works~\cite{wang2021each, lin2022joint} tackle the view discrepancy by focusing on salient points or local patterns. Furthermore, Transformers~\cite{vaswani2017attention} have been explored for their ability to globally capture contextual image features, potentially enhancing the spatial relationship between cross-view images~\cite{zhao2024transfg, dai2021transformer}. Despite these advancements, the challenges remain: learning-based methods that rely on 2D image pair supervision can still be susceptible to overfitting the training data, which limits their generalization on various regions. Furthermore, the burdens of data annotation and network training still limit the transferability of recent methods.

Although self-supervised learning approaches~\cite{li2024unleashing, li2024learning} aim to eliminate the need for annotated pairs, they still require significant computational resources for training and large collections of unlabeled training data. Some works~\cite{Tzachor2024EffoVPREF, keetha2023anyloc} explore the potential of foundation models for visual place recognition. For instance, AnyLoc~\cite{keetha2023anyloc} introduces a non-data-driven model to address the geo-localization problem. However, it struggles with view discrepancies between UAV and satellite images, lacking transferability to new regions.

\subsection{Novel View Synthesis for Geo-localization}
Image-based geo-localization methods traditionally rely on matching query images directly with reference images in a database. However, it faces challenges when the viewpoints of the query image and the reference images differ significantly. Novel view synthesis techniques have emerged as a promising solution to address this viewpoint discrepancy in two primary ways. The first approach involves using novel view synthesis for data augmentation during the training phase of learning-based models~\cite{moreau2022lens, chen2024leveraging}. By introducing synthetic images with varying viewpoints, the model can learn to generalize better to unseen real-world scenarios. Another approach~\cite{uav2022tian} generates novel views during the inference process to eliminate viewpoint discrepancies, effectively aligning the query and reference images to improve performance. However, this method can only handle the fixed view transformation between the drone and satellite images. 

Several works~\cite{place2015torii, regmi2018cross, regmi2019bridging, uav2022tian, toker2021coming} leverage GANs~\cite{wu2022cross} to generate target views from source ground images, requiring labeled image pairs to provide sufficient data for training the network. Recent works explore the Diffusion models~\cite{ho2020denoising} for their ability to produce high-quality cross-view image generation, including satellite-to-ground~\cite{Sat2Scene2024li, qian2023sat2density, shi2022geometry, Lu_2020_CVPR} and ground-to-BEV~\cite{li2024unleashing, wang2024fine}. Furthermore, recent work~\cite{arrabi2024cross} uses multi-modal foundation models~\cite{ramesh2022hierarchical, Rombach2022stable} to guide image generation through textual descriptions. However, a significant limitation of these deep learning-based image synthesis approaches is their tendency to hallucinate unseen or new areas of the scene. This issue can severely impact the accuracy of geo-localization, as these synthetic images may not accurately represent the true geographic layout, leading to errors in retrieval and matching.

\subsection{Multi-view Reconstruction for Geo-localization} 
Multi-view novel view synthesis is particularly well-suited to the UAV geo-localization problem, as UAVs are capable of capturing multiple views of a ground target or leveraging multi-agent systems to cover larger areas. Traditional approaches~\cite{irschara2009structure, Zhang2020ReferencePG} rely on sparse point clouds generated by tools such as Colmap~\cite{schonberger2016structure, pan2024glomap} to reconstruct the geometric structure of a scene from different viewpoints. These point clouds serve as the foundation for more detailed reconstructions, though they often lack fine details and can be computationally expensive. Recent advances~\cite {zhou2024nerfect, moreau2022lens} have introduced radiance fields with learnable representations for visual localization, significantly improving the quality of scene representation. Neural Radiance Fields (NeRF)~\cite{mildenhall2021nerf} offers an implicit and continuous volumetric representation of 3D scenes, enabling high-fidelity rendering of novel views. NeRF achieves impressive results by modeling scene geometry and appearance through neural networks, leading to photo-realistic reconstructions. However, despite their high-quality outputs, NeRF-based methods typically suffer from slow training and inference times, making real-time applications challenging. In contrast, 3D Gaussian Splatting (3DGS)~\cite{kerbl20233d} adopts an explicit representation, modeling 3D scenes using Gaussian-shaped primitives. This approach allows for highly efficient and real-time rendering, which is particularly advantageous in UAV applications where time-sensitive decision-making is crucial. Several recent works~\cite{jiang20243dgs, jun2024renderable} have utilized 3DGS for map representation and camera pose prediction, matching query images to a pre-built Gaussian map. While effective, these approaches face challenges related to the maintenance and frequent updates of the Gaussian map, which are both storage-intensive and computationally demanding. This limitation reduces the scalability of the map to cover larger areas. Our proposed pipeline addresses these limitations by applying 3DGS for query reconstruction in UAV-based geo-localization, ensuring the scalability and efficiency of the reference map. This enables UAVs to perform geo-localization tasks effectively on a global scale.

\section{Methodology}
\label{sec:method}
\begin{figure*}[ht]
	\centering
	\includegraphics[width=0.9\textwidth]{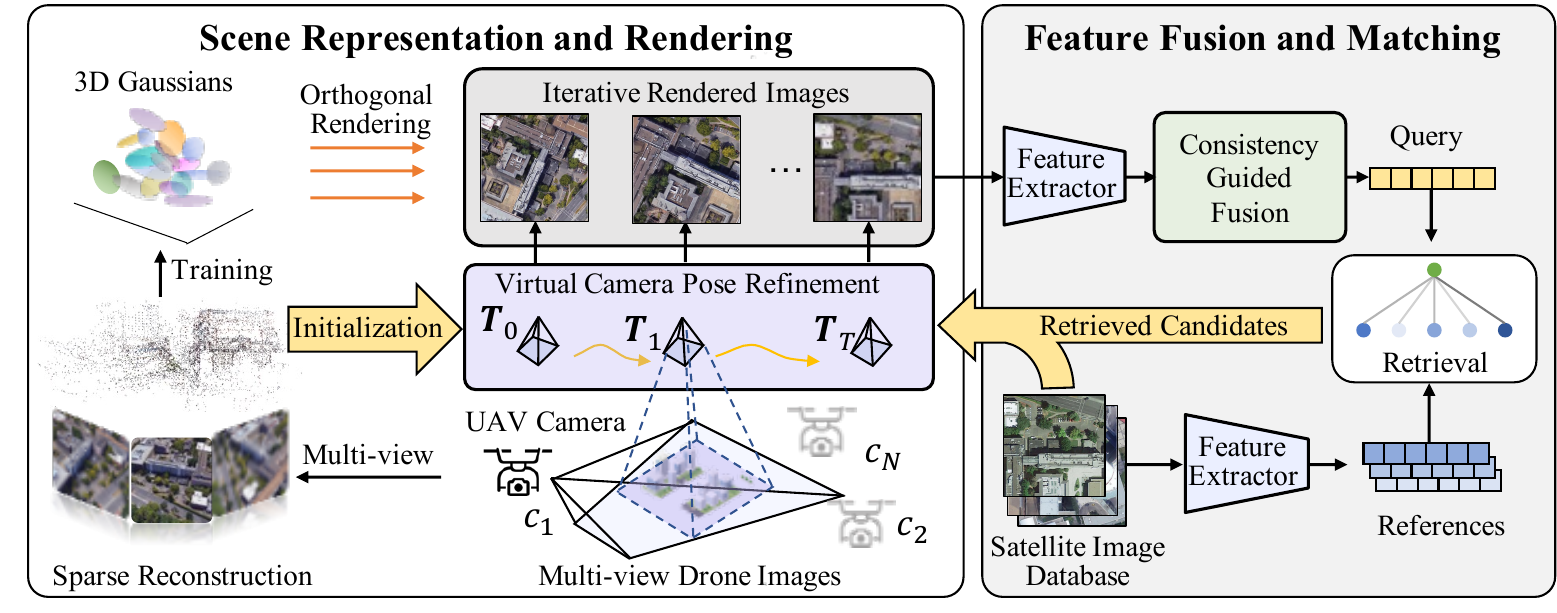}
	\caption{\textbf{Overview of the rendering-based UAV geo-localization.} Given multiple oblique input views, our method first predicts an initial sparse reconstruction and learns a 3D representation using 3DGS. A virtual camera is estimated to render the scene and extract features for matching with real satellite images. The virtual camera pose is then iteratively updated based on feature matching, enabling high-fidelity novel view synthesis that aligns with the true satellite image.}
	\label{fig:pipeline}
\end{figure*}

\subsection{Rendering-based Scene Geo-localization}
\subsubsection{Pipeline Overview}
Given a set of oblique UAV images of a scene, $\mathcal{X}=\{X_1, X_2, ..., X_{N_v} \}$, captured by the drone's camera ($N_v$ denotes the number of drone images), the objective is to match this scene against a database of geo-tagged satellite images, $\mathcal{Y}=\{Y_1, Y_2, ..., Y_M \}$ ($M$ denotes the size of satellite image database). Our approach leverages an iterative scene rendering and retrieval process to achieve it. We first reconstruct a sparse 3D scene from the UAV images to estimate an initial virtual satellite camera pose and initialize 3D Gaussian primitives. The Gaussian Splatting model is then optimized to represent the scene, allowing us to render a virtual image from the virtual camera pose, reducing the view disparity between the UAV and satellite perspectives. Global features are extracted from the rendered image and satellite image database for retrieval. When the retrieved images can be geometrically verified against the reconstructed scene, we iteratively refine the virtual camera pose to improve the alignment between the rendered and potential true satellite images, improving final retrieval accuracy and efficiency. The overall pipeline is shown in Fig.~\ref{fig:pipeline}. Our paradigm offers a distinct advantage over traditional data-driven methods: iteratively refining the retrieval process without requiring pre-training or fine-tuning on a dataset.

\subsubsection{Multi-view Gaussian Scene Rendering}

We hypothesize that the UAV can access multi-view images from the scene in real-world applications. Therefore, we first establish the scene reconstruction from the given multiple drone images $\mathcal{X}$ via the Struction-from-motion (SfM)~\cite{schonberger2016structure}. This estimates the camera pose of every UAV image and performs the sparse reconstruction to produce the scene's point cloud $\mathcal{P}$. Considering the efficient rendering of the recently advanced 3DGS, we adopt the 3DGS to represent the scene from the input multi-view images. 3DGS models the scene by the anisotropic 3D Gaussian primitives, embodying the differential attributes of the volumetric representation of the scene. 

Each point $p_i$ in the sparse point cloud $\mathcal{P}$ is assigned as the initial 3D Gaussian primitive, which is parameterized as $G_i(x)$ with 3D covariance $\bm \Sigma$ and mean $\bm \mu$:

\begin{equation}
    G_i(x) = e^{-\frac{1}{2}(x-\bm\mu)^\top\bm{\Sigma}^{-1}(x-\bm\mu)},
\end{equation}
where $\bm{\Sigma}$ is computed as $\bm \Sigma=\bm{RSS}^\top \bm{R}^\top$, denoting the covariance matrix of the 3D Gaussian with spatial scale $\bm S\in \mathbb{R}^3$ and rotation $\bm R\in \mathbb{R}^{3\times 3}$.

The training process of the Gaussian involves projecting the 3D Gaussian $G(x)$ onto the image plane via splatting. A differentiable rasterizer then sorts the Gaussian primitives by their depth relative to the camera and applies rendering to each pixel color $\bm C$ of the image plane:
\begin{equation}
    \bm C = \sum_{i\in\mathcal{N}} \bm c_i \alpha_i \prod_{j=1}^{i-1} (1-\alpha_j),
\end{equation}
where $\mathcal{N}$ is the set of sorted Gaussian primitives overlapping with the pixel, and $\bm c$ and $\alpha$ represent the color and opacity of the Gaussian primitives, respectively.

The training process for the Gaussian requires the oblique images and their camera pose estimated via SfM. While the oblique images are captured with a perspective projection, this projection differs from the orthogonal projection used in real satellite images. To produce a rendered image that closely resembles a satellite view, we introduce orthogonal projection during the rendering process of the virtual camera.

In camera space, the origin of the coordinate system is at the center of the projection, and the projection plane is defined by $x_2 = 1$. The relationship between the camera space $\bm x = (x_0, x_1, x_2)$ and ray space coordinates $\bm u = (u_0, u_1, u_2)$ is given by the mapping $\bm u=m(\bm x)$, which is defined as:

\begin{equation}
\begin{pmatrix}\label{eq:orth_proj}
        u_0\\ u_1 \\ u_2
    \end{pmatrix} = m(\bm x) =
    \begin{pmatrix}
        2x_0/ s_w \\
        2x_1 / s_h \\
         1 
     \end{pmatrix},
\end{equation}
where $s_w$ and $s_h$ are the adjustable scale factors of width and height of the rendered area, which are determined in Eq.~\ref{eq:adaptive_w}.

This mapping shows that the camera space and the ray space are only related to scale transformation. The Jacobian matrix $\bm J_o$ of orthogonal splatting is given by:
\begin{equation}
\bm J_o = \frac{\partial m(\bm x)}{\partial \bm x} = 
    \begin{pmatrix}
         2/ s_w & 0 & 0 \\
         0 & 2/s_h & 0 \\
         0 & 0 & 0 
    \end{pmatrix}.
\end{equation}
The covariance matrix $\bm \Sigma^{2\mathrm{D}}$ of the projected 2D Gaussian on the image plane can be expressed by:
\begin{equation}
    \bm \Sigma^{2\mathrm{D}} = \bm J_o \bm{W\Sigma W^\top} \bm J^\top_o,
\end{equation}
where $\bm W$ denotes the camera pose for view transformation.

With the orthogonal projection controlled by the scaling factors, the rendered image can neglect the impact of depth variance $x_2$ of the virtual camera, ensuring stable rendering and subsequent update of the virtual camera. 

\subsubsection{Virtual Camera Initialization}
The initial virtual camera $\bm{T}_0 = \{\bm R_0, \bm t_0\}$ is derived from the trained Gaussian primitive cloud and resembles the overhead satellite pose.
To achieve a top-down view of the scene, the virtual camera poses are aligned with an estimated ground plane.
We estimate the ground plane in the point cloud using the RANSAC~\cite{fischler1981random} algorithm, which detects the major plane as the ground within the scene. The camera rotation $\bm{R}_0$ is then defined to align its viewing direction perpendicular to this ground plane. Since the virtual satellite camera employs orthogonal projection, its translation matrix $\bm{t}_0=(t_x,t_y,t_z)^\top$ only needs to consider the $t_x$ and $t_y$ components. In addition, scene scaling factors $s_w$ and $s_h$ are proposed to control view field scaling. We also observed that areas without projected points in the image can negatively impact the final rendering quality. To address this, we designed an optimization objective that minimizes blank areas while maintaining maximum rendering area, resulting in high-quality renderings. The objective is shown below:
\begin{equation}\label{eq:adaptive_w}
    \max_{t_x, t_y, s_w, s_h} \frac{\iint_S H(x, y) - \lambda_m e^{-H(x, y)} \, dx \, dy}{\iint_S H(x, y)  \, dx \, dy},
\end{equation}
where $S = [t_x + \frac{s_w}{2}, t_x - \frac{s_w}{2}, t_y + \frac{s_h}{2}, t_y - \frac{s_h}{2}]$ is the utilized rendering area, $H(x,y)$ is the histogram of the projected points on the ground plane grid $f(x)$, and $\lambda_m$ is the weight to control the impact of the blank point term. With adaptive scaling of the rendered image, the rendered image can retain sufficient scene information and also provide clear rendering performance.

\subsubsection{Initial Global Representation of Scene}
Following recent deep image geo-localization methods~\cite{keetha2023anyloc, zheng2020university}, we employ a pre-trained task-irrelevant foundation model as the feature extractor to encode both the scene and the reference satellite image database into a unified embedding space. The foundation model produces the feature map of each input image, and then the aggregation method (e.g., GeM~\cite{radenovic2018fine}) processes it into a global feature.  The global features $\bm r$ extracted from the satellite image database serve as references $\mathcal{R}=\{\bm r_1, \bm  r_2, ..., \bm r_M\}$. Retrieval is accomplished by ranking the global features based on the cosine similarity of the query and reference.

To enhance the initial global feature $\bm e_0$ of the scene, we combine features from both the original drone images and the rendered images. Considering that the quality of the rendered image can affect its effectiveness in representing the scene, we employ a weighted selection mechanism. We randomly select one of the original oblique drone images and extract its feature vector as $\bm f_0$. The rendered image's global feature is denoted as $\bm f_r$. The similarity between $\bm f_r$ and $\bm f_0$ is computed to determine the weight assigned to the rendered image feature. If the similarity between $\bm f_r$ and $\bm f_0$ is high, the rendered image is considered reliable and contributes more significantly to the final feature representation. Conversely, if the similarity is low, the rendered image's influence is reduced. This weighted aggregation helps mitigate the impact of potential distortions or artifacts in the rendered image. The formulation of $\bm e_0$ is shown below:
\begin{equation}
    \bm e_0 =  <\bm f_0, \bm f_r>   \bm f_r + (1 - <\bm f_0, \bm f_r> ) \bm f_0,
\end{equation}
where the $<,>$ denotes the cosine similarity of the features, this ensures that even if the rendered image is severely distorted, $\bm e_0$ still retains reliable information for initial retrieval. 

\subsection{Iterative Refinement of Rendered Images}
While the rendered images from the initial virtual camera pose resemble satellite images, the initial retrieval may not be fully accurate due to the spatial offset between the rendered image and the real satellite patch. Therefore, we propose an iterative refinement process to improve alignment between the query scene and satellite views. This iterative alignment involves two components:  the iterative update of the camera pose to progressively align the visual appearance of the scene with the satellite images, and a feature fusion that refines the alignment between the scene and satellite features in the embedding space.

\subsubsection{Candidate Camera Pose Update}
\begin{figure}[t]
	\centering
	\includegraphics[width=0.4\textwidth]{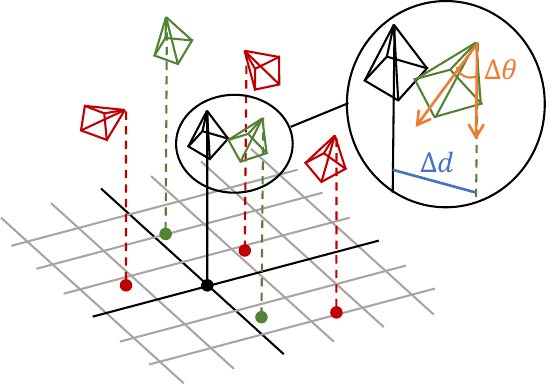}
	\caption{\textbf{Candidate camera pose selection.} The black denotes the previous camera’s coordinate system. Inlier camera poses are marked in green, while outliers are marked in red. In the previous camera’s coordinate, $\Delta d$ is the x-y distance to the previous camera, and $\Delta \theta$, is the angular deviation relative to the z-axis in the previous camera.}
	\label{fig:filter}
\end{figure}

We progressive update of the virtual camera view with the retrieved candidates. The Top-K high similarity features of the retrieved satellite images are selected as candidates to perform feature matching and relative camera pose estimation from the reconstruction scene. 
Considering that only the satellite image covers the scene is helpful for the pose refinement, we filter the outlier candidates using a) matching verification and b) pose verification, and finally perform c) candidate camera update:

a) \textbf{Matching Verification}: The different scenes have different content on the image, leading to a low overall number of matched inliers, we only preserve candidate satellite images with the number of matched feature pair inliers larger than a threshold $N_m$ for subsequent pose estimation.

b) \textbf{Pose Verification}: To ensure appropriate camera candidates for the scene, we perform geometric verification on the candidate camera poses, as shown in Fig.~\ref{fig:filter}. We retain candidates with positions and viewing directions that are similar to those of the previous virtual camera. This verification process involves two key metrics: $\Delta d$ and $\Delta \theta$. First, we evaluate the position offset, $\Delta d$, which reflects the distance within the x-y plane relative to the previous camera's coordinates. Second, we calculate the angular offset, $\Delta d$, between the viewing directions of the current and previous cameras. This is done by computing the rotation angles of the z-axis direction of the candidate cameras relative to the corresponding vector in the previous camera's coordinate system. Candidates with smaller $\Delta d$ and $\Delta \theta$ are considered more likely to represent the true camera pose. 

c) \textbf{Camera Pose Interpolation}: The update of the $k$-th candidate's virtual camera pose at iteration $t$ is performed within the SE(3) space, which represents the group of rigid transformations, combining both rotation and translation. This interpolation is performed between the previous camera pose at iteration $t-1$, denoted by $\bm{T}_{t-1} = \{\bm{R}_{t-1}, \bm{t}_{t-1} \}$, and the filtered candidate pose $\bm{T}_{t, k} = \{\bm{R}_{t, k}, \bm{t}_{t, k} \}$.

\begin{equation}\label{eq:cam_interpo}
\begin{array}{rl}
    \overline{\bm R}_{t, k} &= (\bm{R}_{t, k}\bm{R}_{t-1}^\top)^a \bm{R}_{t-1}, \\ 
    \overline{\bm t}_{t, k} &= \overline{\bm R}_{t, k}((1-a)\bm{R}_{t-1}^\top \bm{t}_{t-1} + a\bm{R}^\top_{t, k}\bm{t}_{t, k}),
\end{array}
\end{equation}
where $a$ represents the interpolation weight for updating the pose. Each filtered candidate generates an updated pose $\overline{\bm T}_{t, k} = \{\overline{\bm R}_{t, k}, \overline{\bm t}_{t, k}\}$ for rendering new images from the trained Gaussian model. The global feature of these rendered image candidates $\bm e_{t, k}$ is then extracted through the feature extractor.

\subsubsection{View Consistency-guided Feature Fusion}
\begin{figure}[ht]
	\centering
	\includegraphics[width=0.45\textwidth]{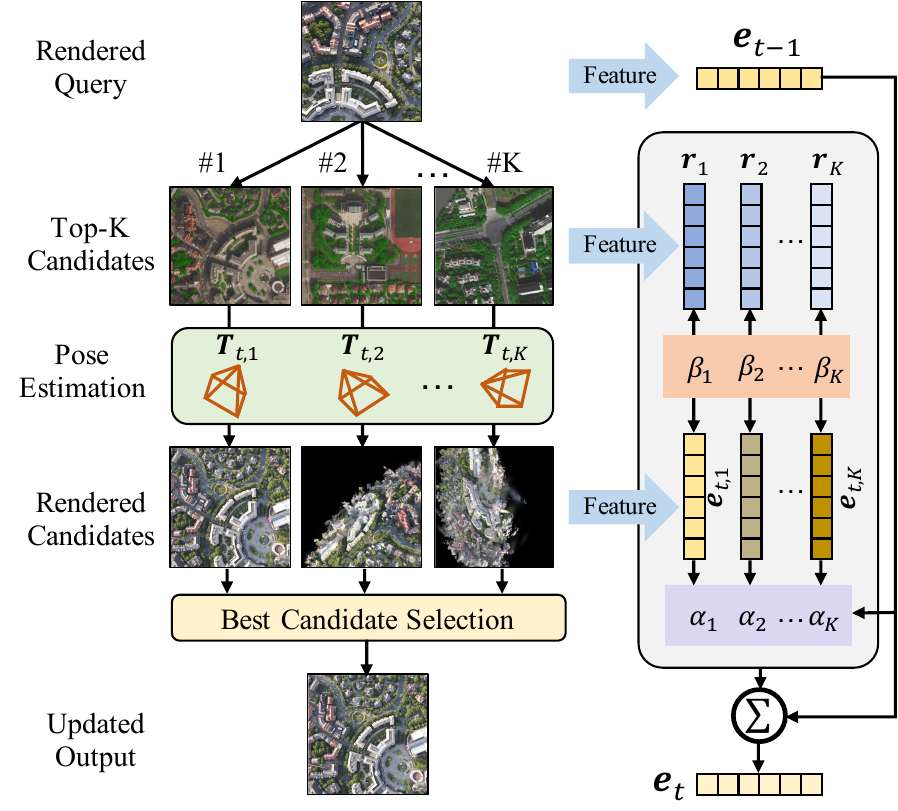}
	\caption{\textbf{The update of the rendered images.}  After extracting the global features of the rendered candidates, we refine the scene’s features using a view consistency fusion module. The module first computes the self-view consistency $\alpha$ between the rendered candidates and the previously rendered view and then calculates the cross-view consistency $\beta$ between the rendered candidates and their corresponding satellite images. The symbol $\sum$ denotes the feature fusion in Eq.\ref{eq:fusion}.}
	\label{fig:iteration}
\end{figure}

After generating multiple filtered rendered images for the scene, the feature fusion can combine the information of these features to retrieve the corresponding satellite images. However, when the retrieved candidates used for updating the camera pose are incorrect, the resulting rendered images may become misaligned with the actual satellite view. Simply combining these global features would introduce noise into the representation. To overcome this, we introduce a consistency-guided fusion module that selectively merges features while preserving alignment accuracy. The fusion module is illustrated in Fig.~\ref{fig:iteration}. 

The fusion module incorporates \textbf{self-view consistency} and \textbf{cross-view consistency}. Self-view consistency ensures that updated rendered images consistently represent the same scene over iteration. Cross-view consistency maintains consistency between the query and retrieved features, ensuring that rendered images from the correct camera pose align with the corresponding satellite image. 

Specifically, the consistency of self-view measures the similarity $\alpha$ between each feature $\bm e_{t, k}$ of the currently rendered image and the feature $\bm e_{t-1}$ of the previously rendered image. The consistency of cross-view calculates the similarity $\beta$ between each $\bm e_{t, k}$ and the feature $\bm r_k$ of its corresponding image from the satellite gallery. Since the current render images are based on the camera poses from the satellite gallery images, the appropriate rendered image should share the same appearance as the corresponding gallery image. The formulations are shown below:

\begin{equation}
\begin{array}{rcl}
    \alpha_k&=&\bm e_{t, k}^\top  \bm e_{t-1},\\
     \beta_k&=&\bm e_{t, k}^\top \bm r_{k}.
\end{array}
\end{equation}

When the query image aligns with the corresponding satellite image, the consistency measures will ideally converge to their maximum values over successive iterations.

The final formulation considers both self-view consistency and cross-view consistency to fuse the candidate query features. We also retain information from the previously rendered image $\bm e_{t-1}$ to regular the feature updation:
\begin{equation}\label{eq:fusion}
    \bm e_{t} = \lambda_s \bm e_{t-1} + (1-\lambda_s) \sum^K_{k=1}  \text{softmax}_k ( \alpha_k \beta_k) \bm e_{t, k},
\end{equation}
where $\lambda_s$ is the weight balancing the new and previous features, and $K$ is the number of filtered candidate features.

We select the best camera pose and its rendered image with the Top-1 ranked similarity $\alpha \times \beta$  for camera pose and feature update of the next iteration. This ensures that the subsequent iteration benefits from the most consistent and aligned rendered image.

\section{Experiments}
\label{sec:experiment}
To evaluate the effectiveness of the proposed pipeline for UAV-to-satellite geo-localization, we conducted a series of experiments on two benchmark datasets, University-1652~\cite{zheng2020university} and SUES-200~\cite{zhu2023sues}. These experiments assess the performance of our method in terms of retrieval accuracy, generalization across datasets, and the impact of various components such as virtual camera pose refinement and the consistency-guided feature fusion module.

\subsection{Dataset and Evaluation Metric}
\textbf{University-1652}:
The University-1652~\cite{zheng2020university} dataset is the first drone-to-satellite geo-localization dataset, providing multi-view drone oblique images and corresponding satellite patches. The dataset comprises images from 1652 university buildings worldwide, where each building includes 1 satellite-view image and 54 drone-view images. The images of 701 buildings are used for network training, while other buildings are reserved for evaluation. 

\textbf{SUES-200}:
The SUES-200 dataset~\cite{zhu2023sues} consists of 200 scenes featuring various university buildings, offering multi-view drone oblique images alongside corresponding satellite images. Each scene includes drone images captured from four different heights, with 50 images per height, providing a comprehensive view of the buildings from diverse perspectives. This dataset is specifically used in our evaluation phase to test the generalization ability of the proposed pipeline, serving as a challenging benchmark for drone-to-satellite geo-localization due to its varied perspectives and heights.

\textbf{Evaluation Metrics}:
To evaluate geo-localization performance, we employ Recall@K (R@K) and Average Precision (AP). R@K measures the proportion of queries where the correct label is among the top-K retrieved results. AP calculates the area under the precision-recall curve, considering the ranking of all positive images. While R@K is sensitive to fixed rankings, AP provides a more comprehensive evaluation.

\subsection{Implementation Details}
For every scene reconstruction, we apply the Colmap~\cite{schonberger2016structure} for camera pose estimation and sparse point initialization. We train the vanilla 3DGS with 7000 iterations for every scene. The resolution of all the rendered images and the reference is 384$\times$384. We directly employ the frozen foundation model DINOv2 as the feature extractor.  The GeM is employed as the aggregation method to process global features. We set the hyperparameters $a=0.8$ in Eq.~\ref{eq:cam_interpo}, $\lambda_m=100$ in Eq.~\ref{eq:adaptive_w}, and $\lambda_s=0.5$ in Eq.~\ref{eq:fusion}. In the iterative refinement, we select the Top 10 candidates for a subsequent camera update and set $N_m = 50$.  We use a single NVIDIA RTX 2080 Ti for all the experiments.

\subsection{Benchmarking Results}

\begin{table*}[ht]
\centering
\caption{Benchmarking Results  on University-1652 \& SUES-200}
\label{tab:compare-u1652}
\begin{tblr}{
cells = {c},
column{2-11} = {22pt},
cell{1}{1} = {r=3}{},
cell{1}{2} = {c=6}{},
cell{1}{8} = {c=4}{},
cell{2}{2} = {c=3}{},
cell{2}{5} = {c=3}{},
cell{2}{8} = {c=2}{},
cell{2}{10} = {c=2}{},
vline{2,8} = {1}{},
vline{2,5,8,10} = {2-7}{},
hline{1,8} = {-}{0.1em},
hline{2} = {2-12}{0.05em},
hline{4} = {-}{0.05em},
}

Methods  &  University-1652 &&&&&& SUES-200 &&& \\
   & $N_v=50$  &  & & $N_v=30$ & & & $N_v=50$~(200m) & & $N_v=50$~(300m) &\\
   & R@1  & R@5 & R@10 & R@1 & R@5  & R@10 & R@1  & AP & R@1  & AP\\
University-1652~\cite{zheng2020university} & \underline{69.33}          & \textbf{86.73}          & \textbf{91.16} & \underline{60.78}          & \textbf{79.37}          & \textbf{86.24}   & \underline{71.25} & \underline{75.81} & \textbf{\textbf{85.63}} & \textbf{\textbf{87.58}}       \\
AnyLoc~\cite{keetha2023anyloc}         & 42.18          & 57.91          & 69.13          & 36.25          & 59.12          & 61.34    & 34.26 & 46.74 & 41.27 & 51.97      \\
Ours (T=0)        & 57.43          & 73.29          &77.14         & 41.28          & 57.64          & 59.13     & 56.00 & 60.50 & 62.00 & 65.00     \\
Ours (T=2)        & \textbf{76.57} & \underline{80.40} & \underline{84.92}          & \textbf{65.18} & \underline{73.28} & \underline{80.34} & \textbf{\textbf{73.00}} & \textbf{\textbf{76.54}} & \underline{76.50} & \underline{77.52}
\end{tblr}
\end{table*}

We evaluate the proposed method in the Drone-to-satellite geo-localization scenario. Since the proposed pipeline requires multiple images of the drone, we averagely select $N_v$ images from the drone image sequence. For a fair comparison, the other methods take the same image sequence and calculate the average values of their global features as the final global feature for retrieval. AnyLoc~\cite{keetha2023anyloc} is the SOTA method that applies an unsupervised model with manually selected layers for general geo-localization tasks. The baseline is trained using the labeled training set of the University-1652 and SUES-200 datasets, respectively. AnyLoc and Our proposed method perform direct inference without any re-training. All methods are evaluated on the test set, with results summarized in Tab.~\ref{tab:compare-u1652}. The best result is highlighted in \textbf{bold} while the second one is highlighted in \underline{underline}. 

\textbf{University-1652}:
The experimental results demonstrate that our method has a substantial improvement in retrieval performance, particularly in Top-1 recall. With a 76\% Top-1 recall, our method outperforms AnyLoc (42\%) by 30 points, largely due to the effectiveness of the rendering technique in reducing the view disparity between UAV and satellite images. Furthermore, without the need for task-specific fine-tuning, our pipeline achieves a 7-point increase in Top-1 recall compared to the supervised method (69\%), highlighting its robust generalization capabilities. In terms of Top-10 recall, we maintain a strong performance close to fine-tuned models. This balance between accuracy and transferability demonstrates that our approach significantly reduces the reliance on extensive data collection and fine-tuning on the task-specific task. 

\textbf{SUES-200}:
Although the supervised method achieves top performance in 300-meter heights, it relies heavily on fine-tuning to learn the region-specific priors. In contrast, our pipeline delivers competitive Top-1 results with 73\% in 200-meter height and 76\% in 300-meter height, all without fine-tuning. We also outperform the method that is not fine-tuned in the same dataset, demonstrating its robustness and efficiency in new regions.

\begin{table}[t]
\centering
\caption{Comparison between recent methods and our proposed method when directly generalized on SUES-200.}
\label{tab:compare-sues}
\begin{tblr}{
  cells = {c},
  column{2} = {10pt},
  column{4-7} = {15pt},
  cell{1}{1} = {r=2}{},
  cell{1}{2} = {r=2}{},
  cell{1}{3} = {r=2}{},
  cell{1}{4} = {c=2}{},
  cell{1}{6} = {c=2}{},
  vline{2,4} = {1-7}{},
  vline{6} = {3-7}{},
  hline{1,3,8} = {-}{},
  hline{2} = {4-7}{},
}
Methods & $N_v$ & Fine-tuning Set & 200m &  & 300m & \\
 &  &  & R@1 & AP & R@1 & AP\\
Zhu~\etal~\cite{zhu2023sues} & 1 & N/A & 13.20 & 17.83 & 14.27 & 18.84\\
Zhu~\etal~\cite{zhu2023sues} & 1 & University-1652 & 63.55 & 68.82 & 72.00 & 76.29\\
AnyLoc & 1 & N/A & 30.15 & 42.57 & 36.74 & 68.62\\
AnyLoc & 50 & N/A & 34.00 & 46.74 & 41.00 & 51.97\\
Ours & 50 & N/A & \textbf{\textbf{73.00}} & \textbf{\textbf{76.54}} & \textbf{76.50} & \textbf{77.52}
\end{tblr}
\end{table}

\subsection{Generalizability Evaluation}
We extend the evaluation to the SUES-200 dataset to assess the generalization ability of our method, which is summarized in Tab.~\ref{tab:compare-sues}. In this experiment, no region-specific data prior to the SUES-200 is provided for training the learning-based methods. While the network without fine-tuning has poor performance, fine-tuning the network with task-specific data, such as University-1652, significantly improves performance. While incorporating multiple frames improves the Anyloc~\cite{keetha2023anyloc} without fine-tuning, it still suffers from view discrepancies, leading to low accuracy. In contrast, our pipeline outperforms these methods without training networks. Compared to directly averaging features of multiple images, exploring structure consistency from multiple input drone images using 3D scene representation proves more effective. This integration of multiple views enhances the scaling and alignment of rendered images, enabling more accurate scene reflection and matching with the target satellite images. These results highlight the robustness and efficiency of our pipeline in new regions and demonstrate its potential for global geo-localization, even without region-specific training.

\subsection{Visualization of Retrieval Results}

\begin{figure*}[ht]
    \centering
    \includegraphics[width=0.9\textwidth]{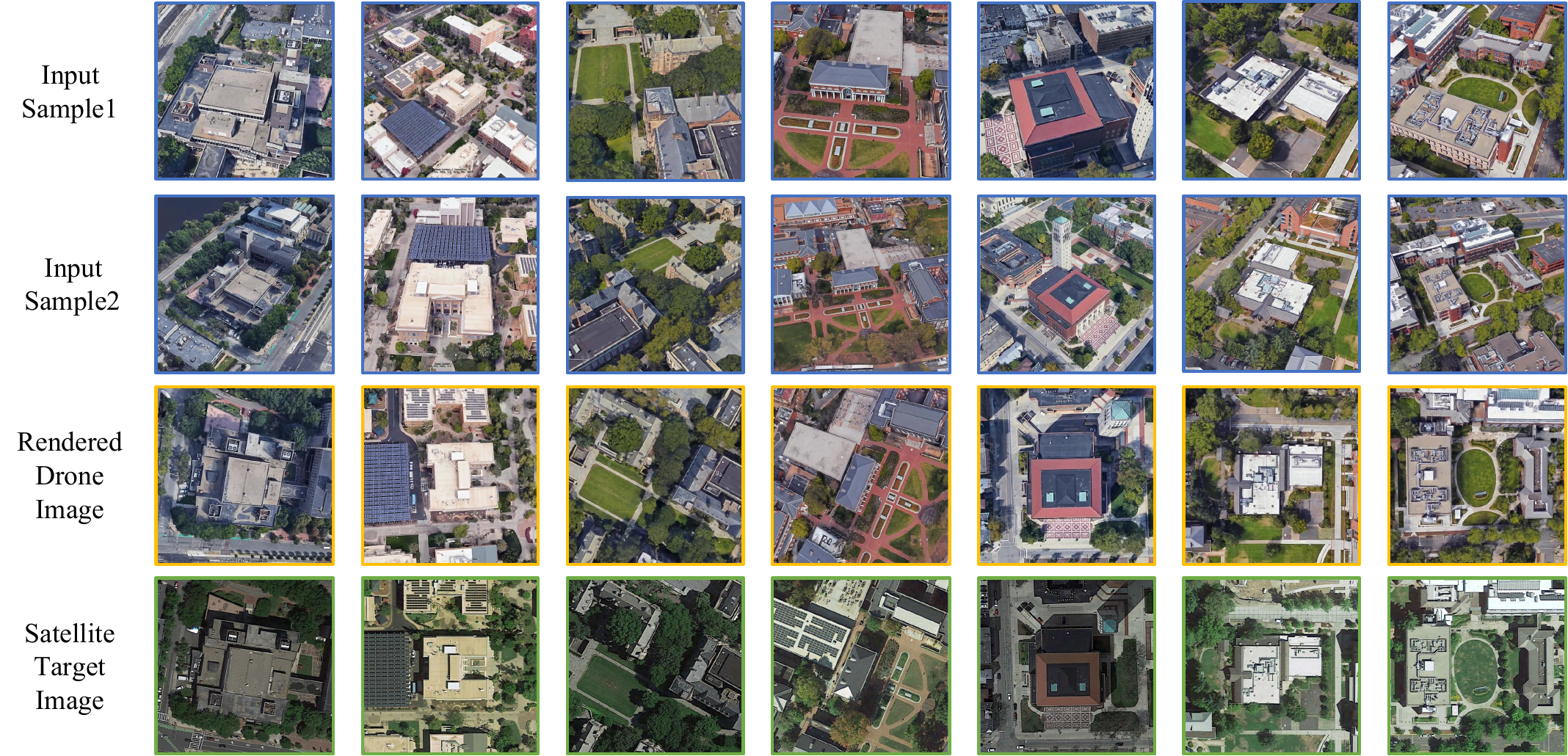}
    \caption{\textbf{Illustration of rendered images and the corresponding true satellite images.} The first two rows show the samples of the input drone images. The third row shows the rendered images of the query scene. The rendered image can align with the true satellite image targets (green boxes).}
    \label{fig:vis}
\end{figure*}

\begin{figure}
\subfigure[Failed rendering and retrieval due to poor camera initialization.]{\includegraphics[width=.45\textwidth]{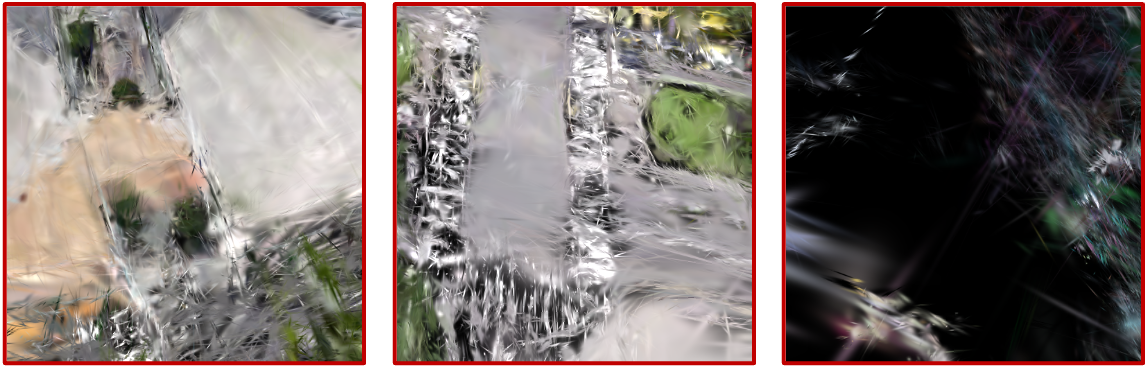}}\\
\subfigure[Successful retrieval despite poor camera initialization.]{\includegraphics[width=.45\textwidth]{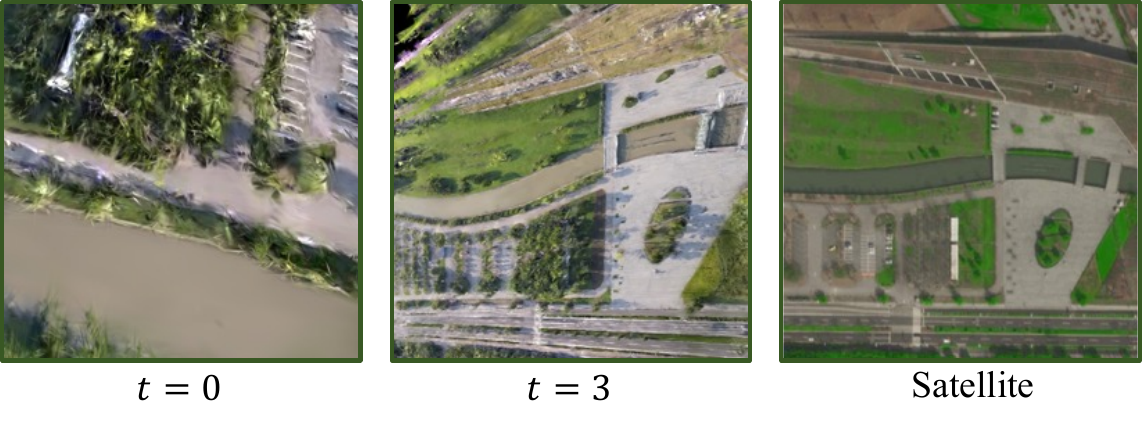}}
\caption{\textbf{The visualization of the failed initialization of rendered images.} The top row depicts that failed initialization of the camera pose leads to distorted rendering and subsequently unsuccessful pose refinement. The bottom row highlights a poor initial camera pose, but after refinement, it successfully aligns to a more accurate pose estimation.}
\label{fig:fail}
\end{figure}

Fig.~\ref{fig:vis} demonstrates the alignment process of the proposed pipeline. Starting with the input samples shown in the first two rows, the rendered images from the virtual camera progressively align with the true satellite view through iterative updates. View discrepancies and viewpoint shifts are effectively eliminated, even in the presence of tonal differences between the true satellite images and the rendered outputs. The proposed pipeline not only aligns the query scene from the drone's perspective but also offers an additional benefit: since updating satellite images for geographic information is often resource-intensive, our pipeline can generate rendered images as potential updates for the satellite gallery, serving as a valuable by-product of our approach.

We examined several failure cases to better understand the limitations of the proposed pipeline, as shown in Fig.~\ref{fig:fail}. Fig.~\ref{fig:fail} (a) shows poorly rendered images, primarily caused by incorrect scale initialization of the rendered scene using Eq.~\ref{eq:adaptive_w} or errors in the initialization of the sparse reconstruction. These issues result in inaccurate feature representations, which in turn lead to invalid updates in the rendered images throughout the iterative process.
Despite these challenges, the pipeline demonstrates some resilience. Even when the scaling of the rendered image is incorrect, we observe that the pipeline can still produce a reasonably reliable scene representation. As shown in Fig.~\ref{fig:fail} (b), with further iterations, the updates help the rendered image gradually align more closely with the true satellite view. These cases underscore the importance of accurate initialization and scaling to ensure the success of the geo-localization process.

\subsection{Ablation Study}\label{sec:ablation}

\textbf{Analysis of Different Modules}:
We evaluate the impact of different modules on the University-1652 dataset, as shown in Tab.~\ref{tab:abla_module}. We use AnyLoc~\cite{keetha2023anyloc} for single-image retrieval, serving as our baseline. For multiple-view input, the average of the image features from the scene is used as the global feature~\cite{zheng2020university, uav2022tian}, which enhances matching accuracy due to the shared scene context. The rendering module further leverages the geometric consistency of multiple images, resulting in better scene representation. Iterative refinement significantly improves retrieval performance by adjusting the camera pose of the rendered image and aligning the query image with the corresponding target. Additionally, the proposed consistency fusion strategy, which incorporates both self-view and cross-view consistency, enhances the feature representation after each update of the rendered image. Therefore, these modules improve the overall performance with iterative refinement.

\begin{table}
\centering
\caption{Ablation study of the proposed modules on University-1652}
\label{tab:abla_module}
\begin{tblr}{
  cells = {c},
  cell{1}{2} = {c=5}{},
  vline{2} = {1-7}{},
  hline{1,8} = {-}{0.1em},
  hline{2,6} = {-}{0.05em},
}
Modules & Configurations &  &  &  & \\
Multiple Views &  & \checkmark & \checkmark & \checkmark & \checkmark\\
Rendering &  &  & \checkmark & \checkmark & \checkmark\\
Iterative Refinement &  &  &  & \checkmark & \checkmark\\
Consistency Fusion &  &  &  &  & \checkmark\\
R@1 & 31.60 & 42.18 & 54.37 & 65.15 & \textbf{76.57}\\
R@10 & 46.52 & 69.13 & 70.19 & 74.12 & \textbf{84.92}
\end{tblr}
\end{table}

\begin{figure}
    \centering
    \begin{tabular}{cccc}
$t=0$ & \includegraphics[width=.2\linewidth,valign=m]{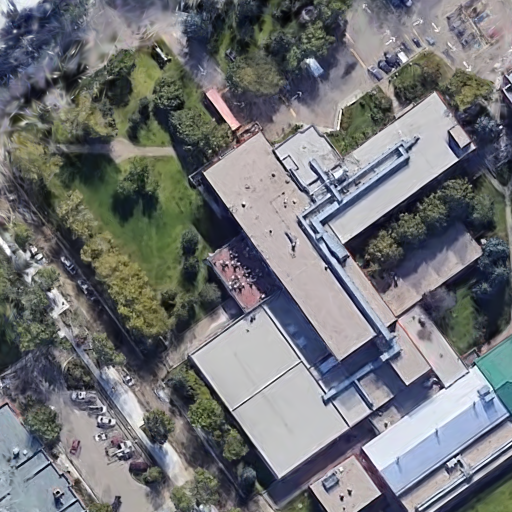} & \includegraphics[width=.2\linewidth,valign=m]{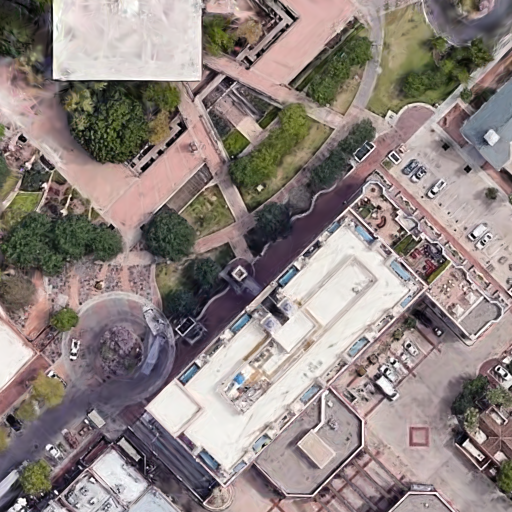} & \includegraphics[width=.2\linewidth,valign=m]{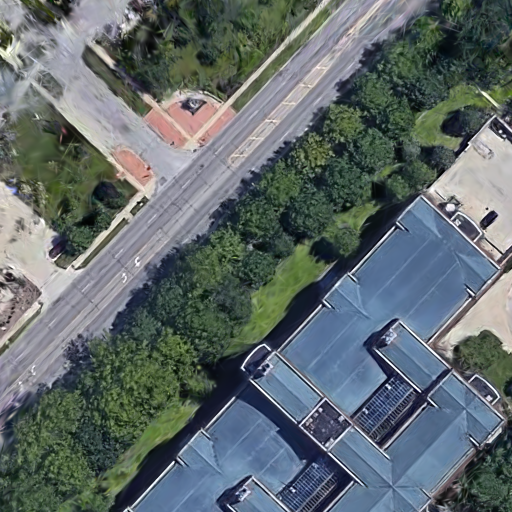}\\[4ex]
$t=1$ & \includegraphics[width=.2\linewidth,valign=m]{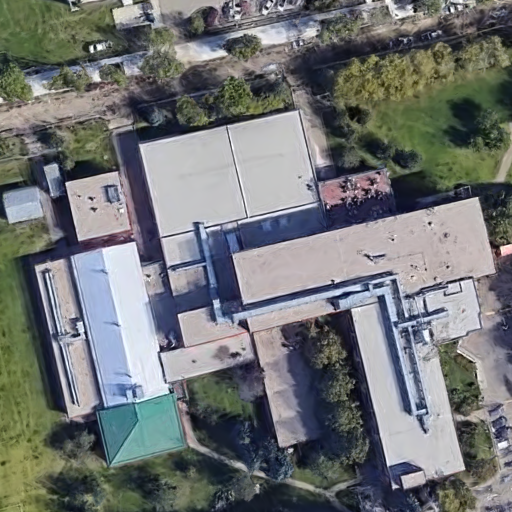} & \includegraphics[width=.2\linewidth,valign=m]{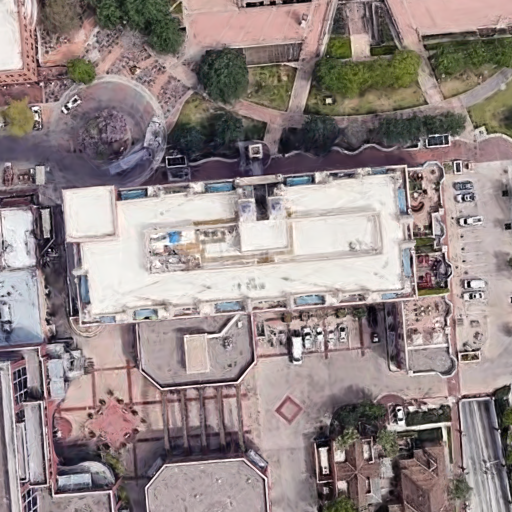} & \includegraphics[width=.2\linewidth,valign=m]{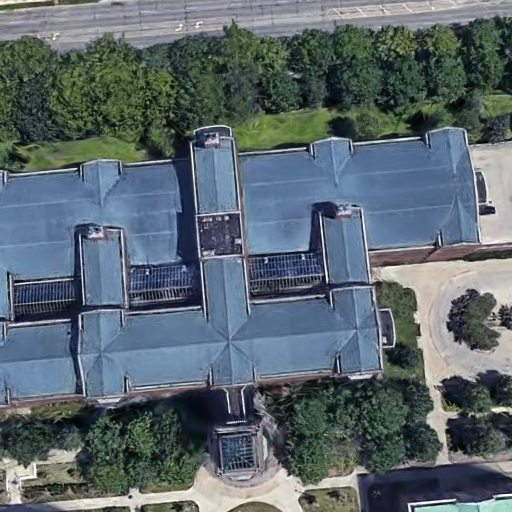}\\[4ex]
$t=2$ & \includegraphics[width=.2\linewidth,valign=m]{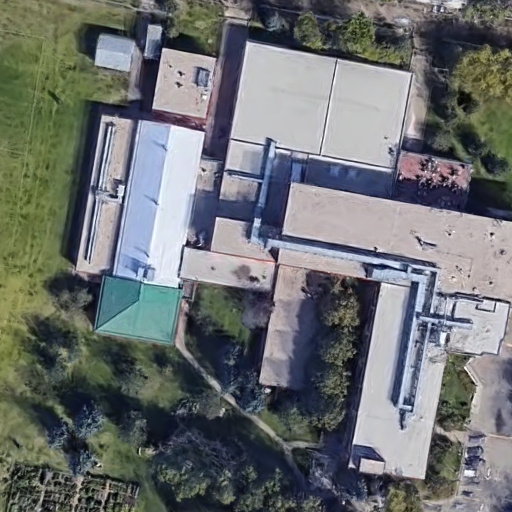} & \includegraphics[width=.2\linewidth,valign=m]{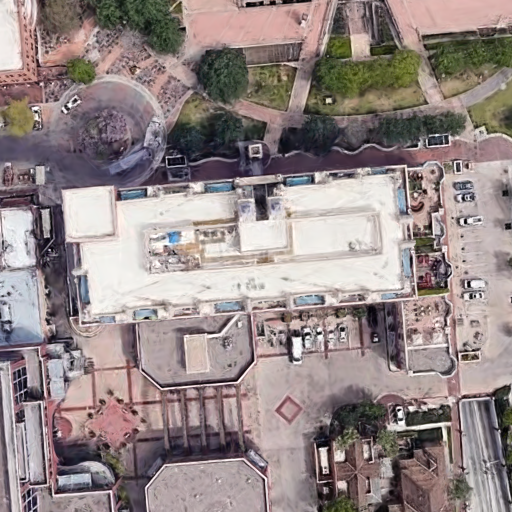} & \includegraphics[width=.2\linewidth,valign=m]{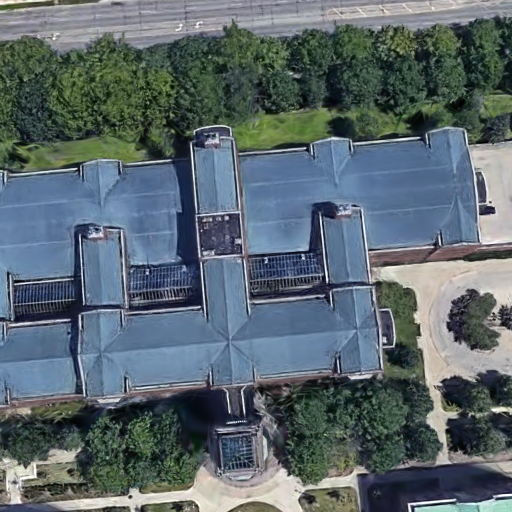}\\[4ex]
$t=3$ & \includegraphics[width=.2\linewidth,valign=m]{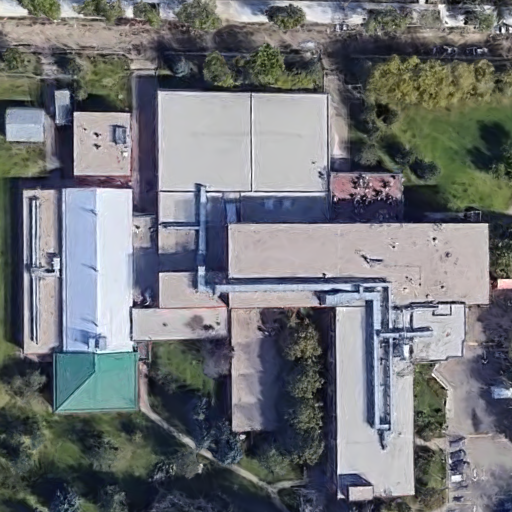} & \includegraphics[width=.2\linewidth,valign=m]{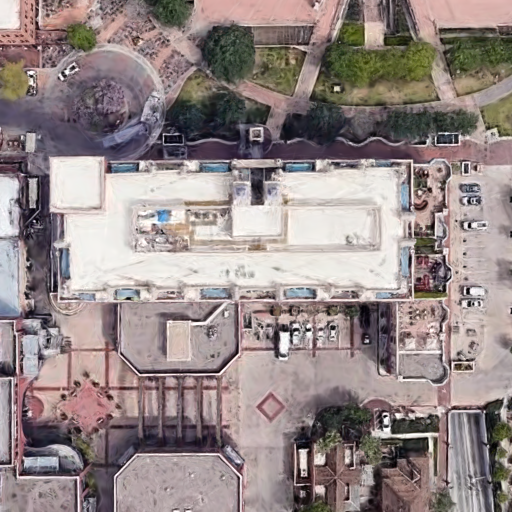} & \includegraphics[width=.2\linewidth,valign=m]{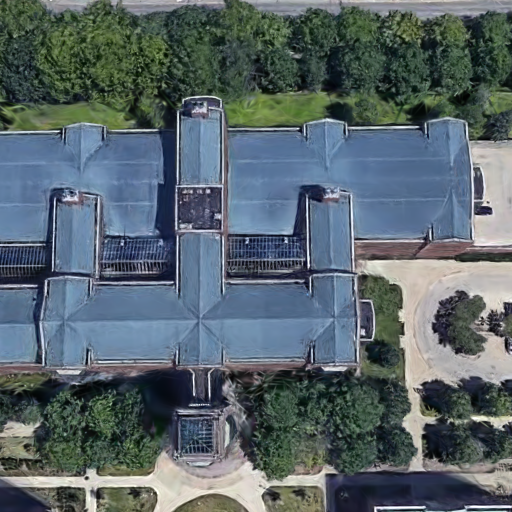}\\[4ex]
Satellite & \includegraphics[width=.2\linewidth,valign=m]{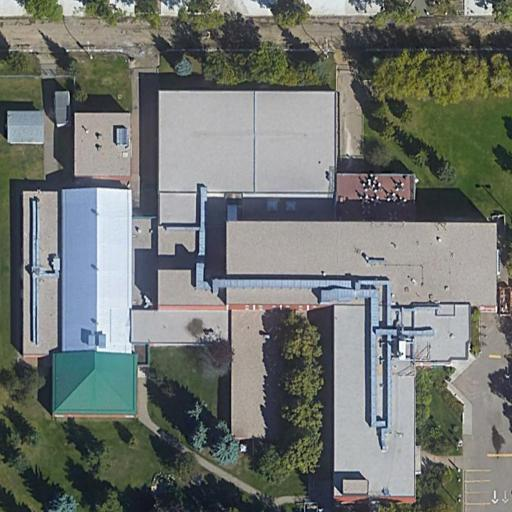} & \includegraphics[width=.2\linewidth,valign=m]{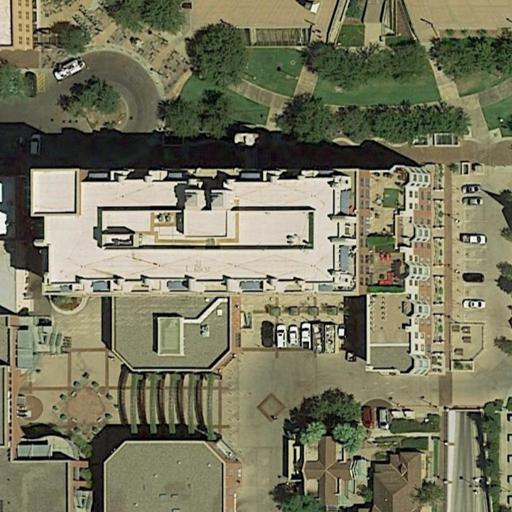} & \includegraphics[width=.2\linewidth,valign=m]{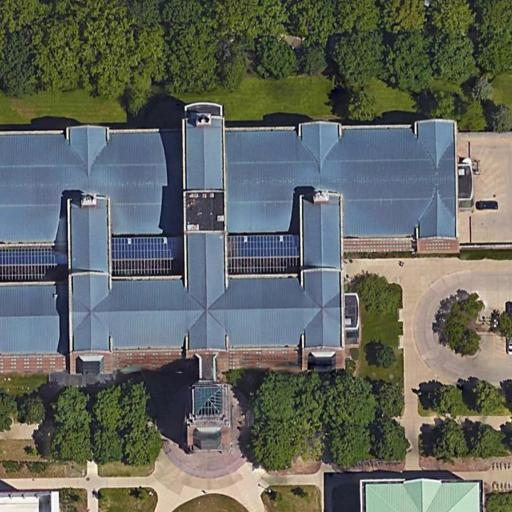}\\
  & (a) & (b) & (c) \\
\end{tabular}
    \caption{\textbf{Rendered images during iterative camera pose update.} The progression across time steps (from t=0 to t=3) demonstrates how the rendered image becomes increasingly aligned with the satellite image}
    \label{fig:iter_render}
\end{figure}

\textbf{Steps of the Iterative Camera Pose Update}:
To provide a comprehensive analysis demonstrating the effectiveness of the proposed iterative process, we performed an ablation study of the number of iteration steps. Tab.~\ref{tab:abla-iter} shows that the increasing overall iteration can improve the matching performance. Especially, the first refinement can largely improve the matching because of the verification using the retrieved candidates. Fig.~\ref{fig:iter_render} demonstrates that when the true satellite image is retrieved among the top-K candidates, the virtual camera pose can be iteratively adjusted to align the rendered image with the satellite view. This iterative process effectively handles cases where the initial rendered images have a large angular offset (Fig.~\ref{fig:iter_render}(a) and (b)) or minimal scene overlap (Fig.~\ref{fig:iter_render}(c)). The results indicate that the alignment stabilizes at $T=2$, achieving accurate matching with the target satellite image.

% \begin{figure}[ht]
%     \centering
%     \includegraphics[width=0.45\textwidth]{fig/iter_img.png}
%     \caption{\textbf{Rendered images with iterative camera pose updation.}}
%     \label{fig:iter_render}
% \end{figure}

% \usepackage{tabularray}
\begin{table}[t]
\centering
\caption{Ablation study of iteration steps for the camera pose update}
\label{tab:abla-iter}
\begin{tblr}{
  column{3} = {c},
  column{4} = {c},
  column{5} = {c},
  column{6} = {c},
  cell{2}{1} = {r=2}{},
  cell{4}{1} = {r=2}{},
  hline{1,6} = {-}{0.1em},
  hline{2,4} = {-}{0.05em},
}
                & T  & 0     & 1     & 2     & 3     \\
University-1652 & R@1 & 57.43 & 63.13 & 76.57 & 76.57 \\
                & R@5 & 73.29 & 78.51 & 80.40 & 79.12 \\
SUES-200        & R@1 & 62.00 & 68.00 & 76.50 & 76.50 \\
                & R@5 & 74.50 & 78.00 & 80.50 & 79.00 
\end{tblr}
\end{table}

\textbf{Impact of the Number of Drone Images}: The proposed pipeline relies on multiple input images to improve query representation and align it with the target satellite image. When only a single image is available, it is fed directly to the feature extractor for retrieval, bypassing the advantages of multi-view input. To analyze the impact of the number of input drone images on geo-localization performance, we conducted experiments on two datasets, as summarized in Tab.~\ref{tab:abla-view}. The results indicate a noticeable decline in performance as the number of images decreases, with a significant drop below 20 images. This is primarily due to the challenges posed by sparse views, which negatively affect the reconstruction and rendering stages. While Colmap produces robust camera pose estimations with fewer views, the sparsity of the reconstructed points hinders the initialization of Gaussian primitives. Additionally, with fewer images available for training the Gaussians, the jumble mess increases, causing the scene—especially buildings—to appear blurred and distorted. Rendering from sparse views using 3D Gaussian Splatting further exacerbates this issue, resulting in blurred or under-sampled regions due to the limited density of Gaussian primitives projected onto the image plane.

\begin{table}[t]
\centering
\caption{Ablation study on the number of drone images}
\label{tab:abla-view}
\begin{tblr}{
  cells = {c},
  cell{1}{2} = {c=2}{},
  cell{1}{4} = {c=2}{},
  vline{2} = {-}{},
  vline{4} = {-}{},
  hline{2-3} = {-}{},
  hline{1,9} = {-}{0.1em},
}
Dataset & University-1652 &  & SUES-200 & \\
$N_v$ & R@1 & R@10 & R@1 & R@10\\
1 & 31.60 & 46.52 & 41.00 & 52.00\\
10 & 47.93 & 66.19 & 53.00 & 59.00\\
20 & 59.53 & 77.83 & 62.50 & 70.00\\
30 & 65.18 & 80.34 & 64.50 & 72.00\\
40 & 72.18 & 81.63 & 73.00 & 75.00\\
50 & 76.57 & 84.92 & 76.50 & 77.50
\end{tblr}
\end{table}

\begin{figure}[t]
\centering
\begin{tabular}{cccc}
$N=9$ & \includegraphics[width=.2\linewidth,valign=m]{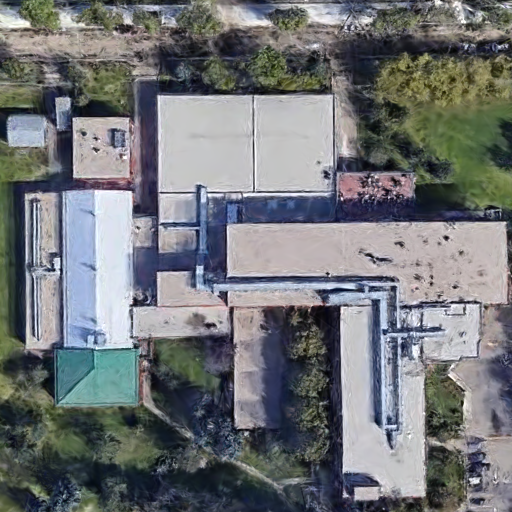} & \includegraphics[width=.2\linewidth,valign=m]{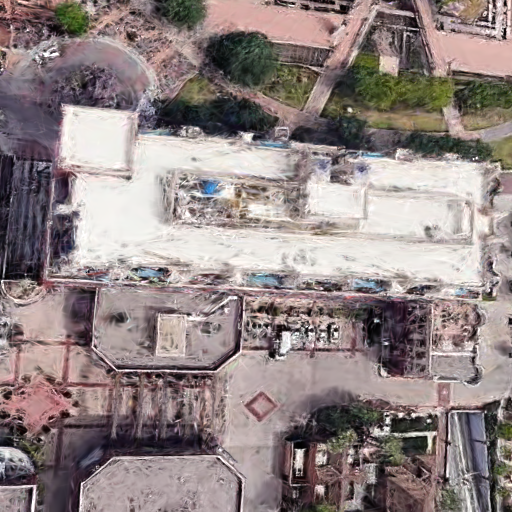} & \includegraphics[width=.2\linewidth,valign=m]{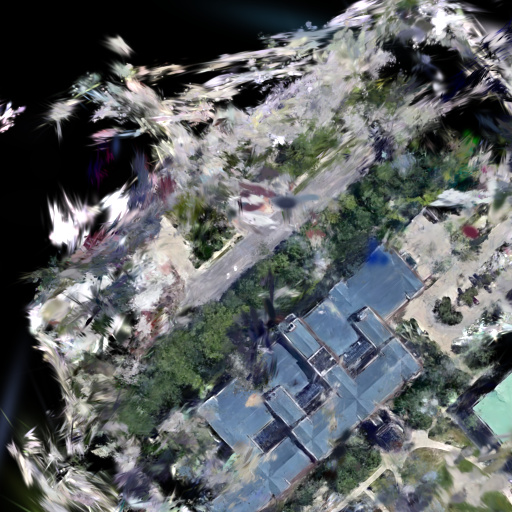}\\[4ex]
$N=18$ & \includegraphics[width=.2\linewidth,valign=m]{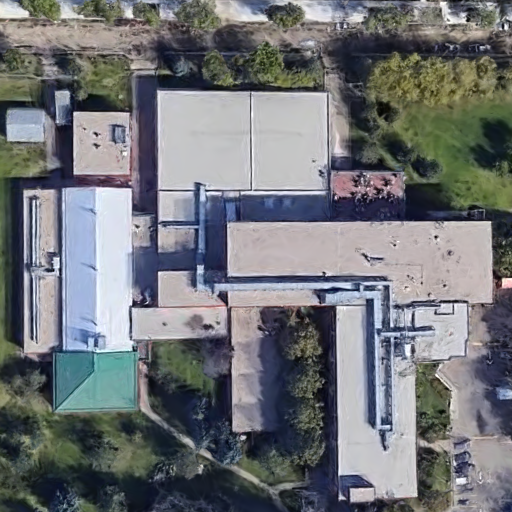} & \includegraphics[width=.2\linewidth,valign=m]{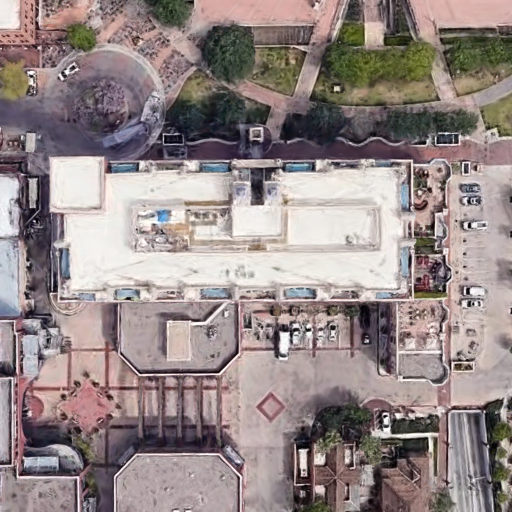} & \includegraphics[width=.2\linewidth,valign=m]{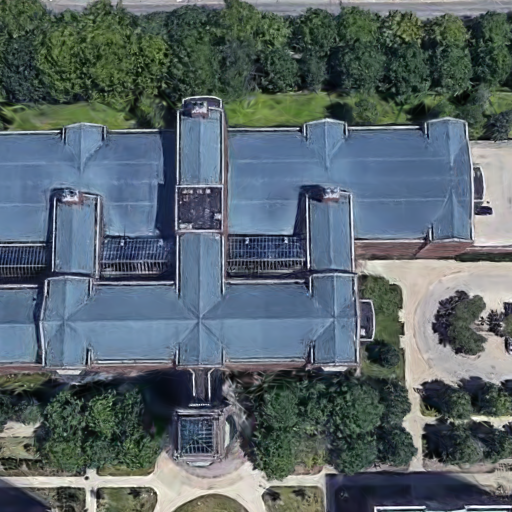}\\[4ex]
$N=50$ & \includegraphics[width=.2\linewidth,valign=m]{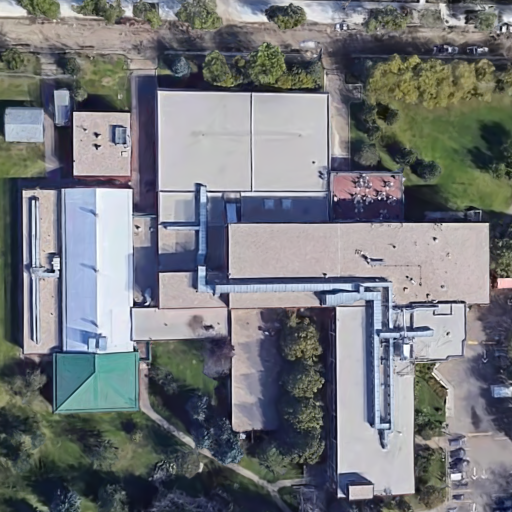} & \includegraphics[width=.2\linewidth,valign=m]{fig/sparse/0010_0010_18.png} & \includegraphics[width=.2\linewidth,valign=m]{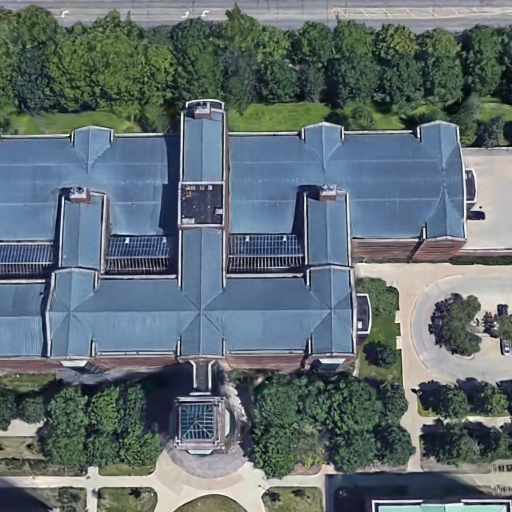}\\[4ex]
Satellite & \includegraphics[width=.2\linewidth,valign=m]{fig/sparse/0000.png} & \includegraphics[width=.2\linewidth,valign=m]{fig/sparse/0010.png} & \includegraphics[width=.2\linewidth,valign=m]{fig/sparse/0707.png}\\
  & (a) & (b) & (c) \\
\end{tabular}
\caption{\textbf{Rendered images with different numbers of the input drone view.} As the number of input views increases, the rendered images become more aligned with the satellite images, resulting in higher fidelity in the novel views.}
\label{fig:sparse_image}
\end{figure}

Fig.~\ref{fig:sparse_image} demonstrates the impact of varying the number of drone images on the quality of the rendered output. As the number of views decreases, the details of the objects in the rendered images become increasingly noisy, which can severely hinder the iterative alignment process with the target image. This noise not only diminishes the visual fidelity of the output but can also lead to misalignment, compromising the effectiveness of geo-localization. However, when the number of views exceeds 18, the rendered images exhibit improved reliability and performance. At this threshold, the quality of the rendering stabilizes and the output is consistent with the target imagery. Although increasing the number of views beyond this point further enhances the rendering details, it is important to note that the geo-localization accuracy tends to reach a saturation point. This indicates that while more views can refine the visual representation, the additional benefits may not translate into significant improvements in localization accuracy.

\begin{figure}[t]
    \centering
    \includegraphics[width=0.48\textwidth]{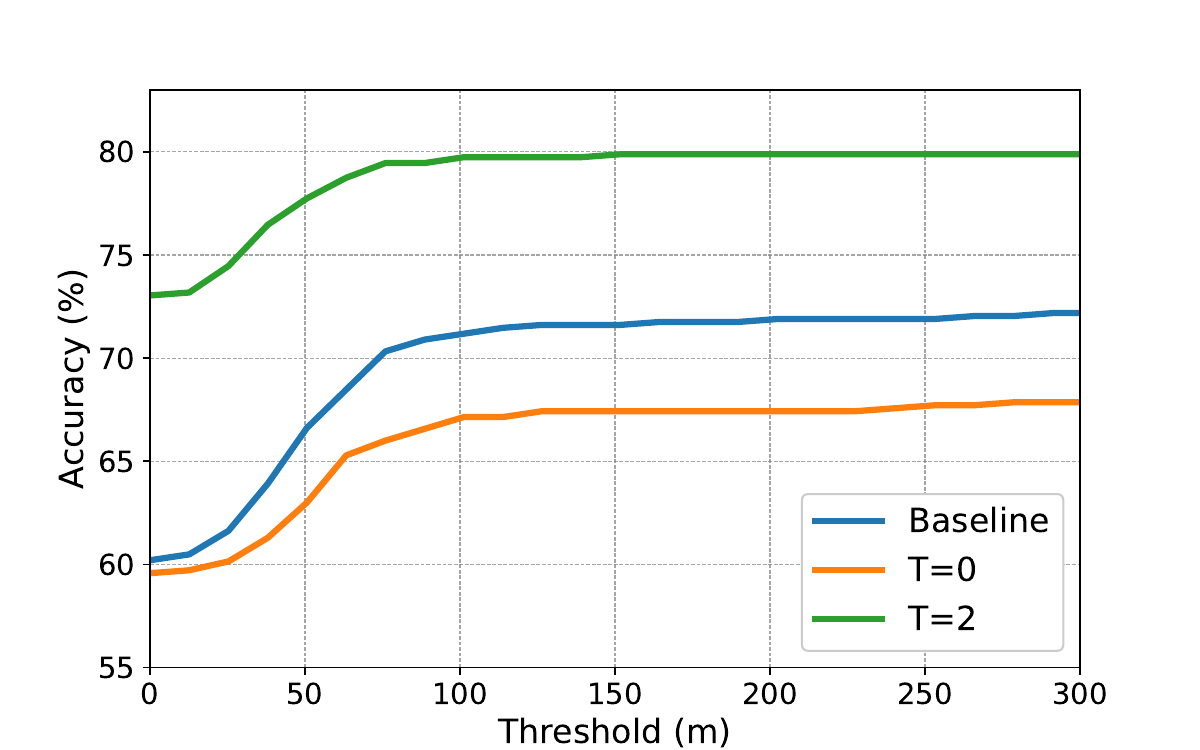}
    \caption{\textbf{Meter-level accuracies in University-1652.}}
    \label{fig:meter}
\end{figure}

\begin{figure}
\centering
\fontsize{8pt}{8pt}\selectfont
\begin{tabular}{cc}
 \includegraphics[width=.3\linewidth,valign=m]{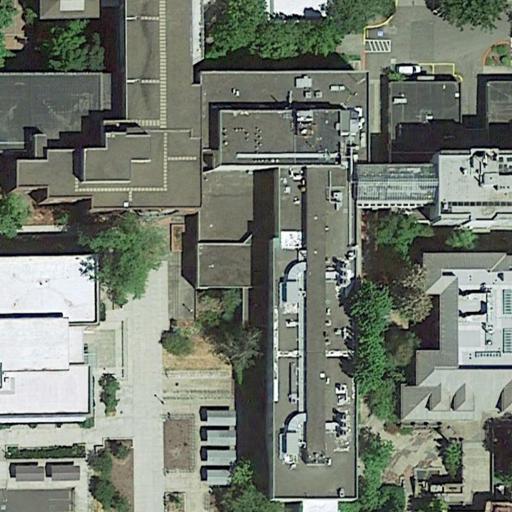} & \includegraphics[width=.3\linewidth,valign=m]{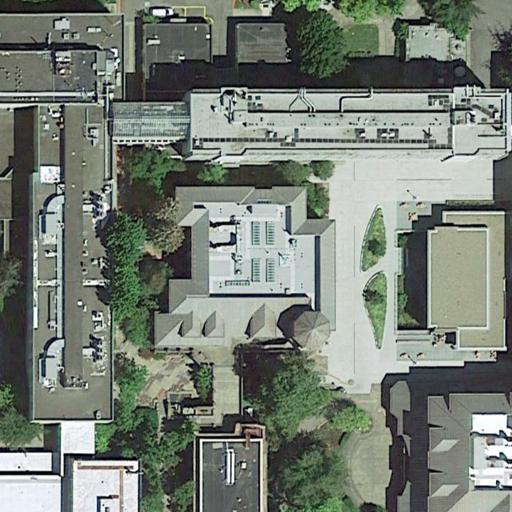} \\ [38pt]
(a) Satellite \#1024 & (b) Satellite \#1018 \\[5pt]
\includegraphics[width=.3\linewidth,valign=m]{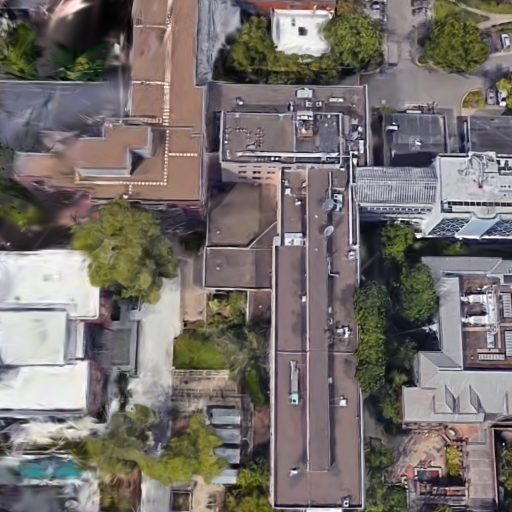} & \includegraphics[width=.3\linewidth,valign=m]{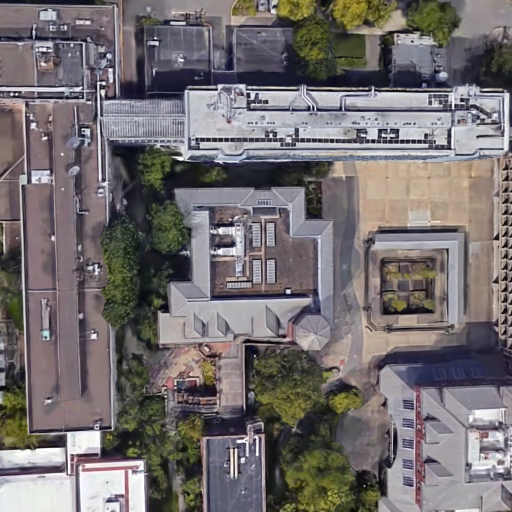} \\ [38pt]
(c) Rendered image of scene&(d) Rendered image of scene\\
 \#1018 aligns target \#1024.&\#1018 aligns target \#1018.\\
\end{tabular}
\caption{\textbf{Uncertainty of the matched target Rendered images in overlap satellite patches.}}
\label{fig:negihbour}
\end{figure}

\textbf{Geographical Comparison}:
To assess the geographical accuracy of the retrieval, we use GPS annotation to measure the meter-level distance between the retrieved target and the true positive target. Fig.~\ref{fig:meter} illustrates the error distribution at the meter level. Due to the overlap in satellite patches, the retrieval may occasionally match a neighboring patch, causing the curve of the baseline method to fluctuate sharply below 100 meters. By incorporating iterative camera pose updates, the rendered image better aligns with the target scene, thereby enhancing the geographical precision. However, a slope persists despite the improvements. This can be attributed to the overlap of the satellite patches, which introduces uncertainty in scene representation. When drone images capture a region covered by multiple satellite patches, the camera pose might mistakenly align with a neighboring patch rather than the labeled target. As shown in Fig.~\ref{fig:negihbour}, (a) and (b) represent two overlapping satellite patches, while (c) and (d) are the rendered images from the same scene. Although this scene is labeled to match and fully covers patch (b), it may still align with its neighboring patch (a), leading to inaccurate metrics. Furthermore, when the virtual camera occasionally aligns with the satellite patch of (a), the rendered image shown in (c) exhibits artifacts and noise due to incomplete scene coverage, which degrades the overall image quality.

\begin{figure}[t]
    \centering
    \includegraphics[width=1\linewidth]{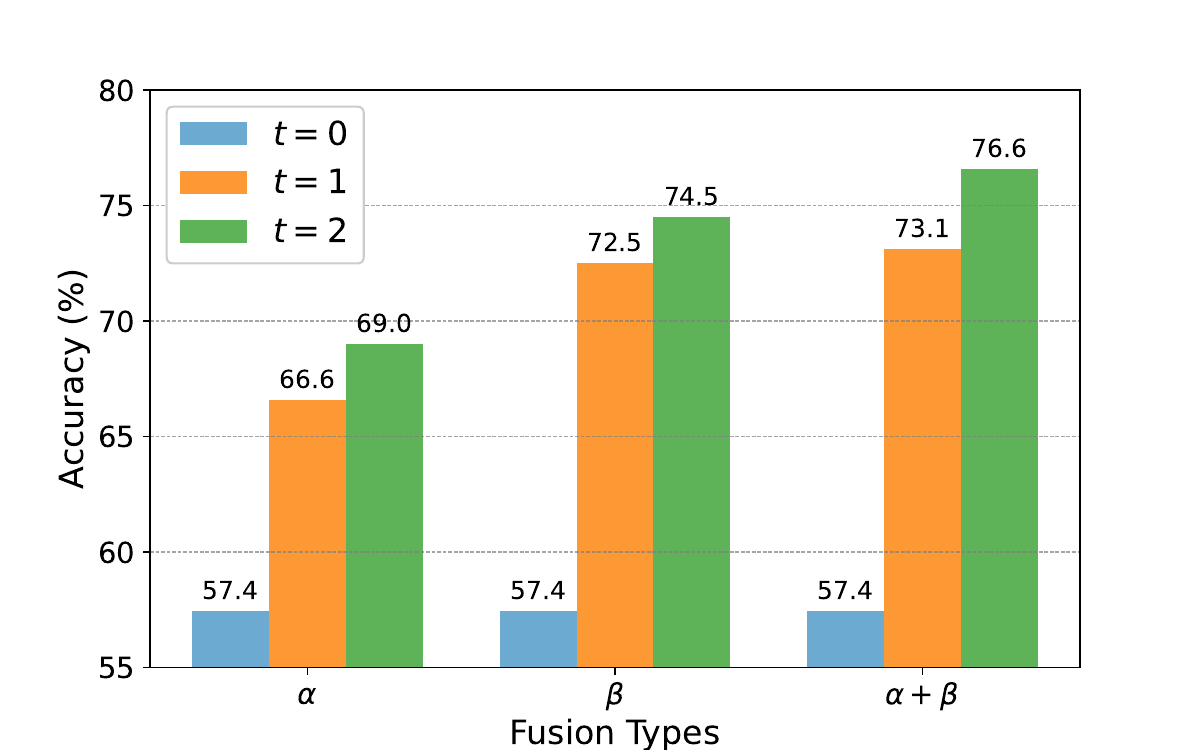}
    \caption{\textbf{Top-1 accuracies of different fusion types.} We perform an exhaustive ablation to report the improvement of different combinations of view consistency regularization on the fusion module. Combining both view consistencies yields the highest top-1 accuracy overall, with improved performance as the iterations progress.} 
    \label{fig:compare-sim}
\end{figure}

\textbf{Impact of Fusion Types}:
The proposed view consistency approach enables the fusion of global features from rendered image candidates based on their similarity to both the previously rendered images and the satellite images used for estimating camera poses. We evaluate three fusion strategies—self-view consistency, cross-view consistency, and their combination—and analyze how each regularization technique improves scene representation. The performance impact of these strategies is illustrated in Fig.~\ref{fig:compare-sim}.

When applying only self-view consistency (denoted as $\alpha$), we observe limited improvement during the refinement process. This regularization ensures that the newly rendered image maintains a similar viewpoint to the previously rendered one, but it does not evaluate the similarity between the rendered image and the retrieved satellite view. As a result, the pipeline exhibits less refinement of the features under this strategy. In contrast, cross-view consistency (denoted as $\beta$) enforces alignment between the rendered images and their corresponding satellite images. Since the rendering camera pose is derived from the satellite view, achieving high cross-view consistency indicates that the rendered image matches the satellite target. This regularization significantly improves the pipeline's ability to align with the correct scene.

Combining both self-view and cross-view consistencies ($\alpha+\beta$) leads to the best performance. This dual consistency ensures that the pipeline maintains coherence across iterations while effectively aligning the rendered images with the target satellite view, leading to enhanced geo-localization accuracy and overall performance.

\subsection{Discussion}

\textbf{Geo-location Accuracy in Image Retrieval}:
A critical challenge in evaluating geo-localization performance lies in reconciling continuous location error with the non-continuous nature of target identification in image retrieval. Accuracy is compromised when the scene aligns with the covered neighboring satellite patch rather than the correct target ID. However, this misalignment on ID results in a significant decrease in retrieval top-K accuracy, even if the geo-location error remains low. As shown in Fig.~\ref{fig:negihbour}, overlapping satellite patches can both be considered positive samples, as the drone image covers them and successfully renders the corresponding image. This observation has motivated us to propose a more refined evaluation metric for future datasets. The new metric would address overlapping tiles by more precisely assessing the exact location of the query scene, focusing on key structures (e.g., buildings) within the scene, and distinguishing between fully and partially covered areas.

\textbf{The Generalization of the Pipeline}:
The pipeline's integration of 3DGS for rendering and scene reconstruction, combined with a task-agnostic foundation model for feature extraction, demonstrates its inherent flexibility. This modular design allows the pipeline to adapt to different scenarios, such as sparse-view or real-time applications, by switching between various techniques. For example, in dense view scenarios, standard 3DGS can be employed to produce efficient and high-quality renderings. In sparse view situations, methods that improve 3DGS for sparse data are equally suitable for integration into the pipeline. This adaptability enhances the generalization capabilities of the pipeline, making it more versatile and effective in real-world UAV geo-localization tasks.

\textbf{Limitations and Future Works}:
 While the proposed rendering-based UAV geo-localization pipeline offers significant advantages, future work could address several limitations. First, the computational cost of processing and reconstructing a 3D scene from multiple images is non-trivial. The current pipeline relies on extensive 3DGS operations, which may degrade time performance, especially when applied to larger scenes or regions. To enhance the real-time capabilities of the pipeline, future research could explore more efficient scene representation techniques or optimization strategies for 3D reconstruction and rendering. Another limitation is the reliance on the initial quality of the 3D reconstruction. Insufficient image coverage or poor initialization of the Gaussian primitives can result in degraded rendered images, leading to inaccurate retrieval and refinement stages. Future work could focus on improving robustness in scenarios with sparse views or incomplete data by incorporating additional sensor data, such as LiDAR or depth information, to obtain more precise scene reconstruction.

\section{Conclusion}
\label{sec:conclusion}
In this paper, we propose a novel pipeline for UAV geo-localization that leverages 3D Gaussian Splatting for efficient scene representation and rendering.
%, coupled with a task-irrelevant foundation model for feature extraction. 
Our approach addresses challenges in cross-view image retrieval by minimizing the view discrepancy between drone-captured images and satellite imagery through high-fidelity rendering. The pipeline further enhances retrieval accuracy by incorporating an iterative camera pose update process, aligning rendered images with their true satellite counterparts, without fine-tuning or data-driven training.
Extensive experiments on the University-1652 and SUES-200 datasets demonstrate the generalizability and robustness of our method in various scenarios. Our pipeline offers a promising solution for UAV geo-localization applications, providing accurate results without the computational overhead of network training or traditional map maintenance.

\section*{Acknowledgment}
The authors would like to express their sincere thanks to the anonymous reviewers for their valuable comments and suggestions. The numerical calculations presented in this article were performed on the supercomputing system at the Supercomputing Center of Wuhan University.

\bibliographystyle{IEEEtran}
\small
\bibliography{ref} %IEEEabrv,

\end{document}